\documentclass[letterpaper]{article} 
\usepackage{aaai2027} 
\usepackage[hyphens]{url} 
\usepackage{graphicx} 
\usepackage{natbib} 
\usepackage{caption} 
\usepackage{amsmath}
\usepackage{amssymb}
\usepackage{booktabs}
\title{Not All Entropy Is Equal: Reward-Conditioned Entropy Allocation in Reinforcement Learning with Verifiable Rewards}
\author{
	Song Yu,
	Li Li\thanks{Corresponding author.},
	Zhenming Wu
}
\affiliations{
	School of Computer and Information Science, Southwest University, Chongqing 400715, China\\
	\texttt{yusong0929@email.swu.edu.cn}, \\
	\texttt{lily@swu.edu.cn},\\
	\texttt{swuwzm666@email.swu.edu.cn}
}
\begin{document}
\nocopyright
\maketitle

\begin{abstract}
Policy entropy is widely used to diagnose exploration and collapse in reinforcement learning with verifiable rewards (RLVR), yet its magnitude does not consistently track reasoning performance across optimization objectives. We argue that this ambiguity arises because global token entropy conflates uncertainty in successful and failed trajectories, diversity across correct solution strategies, and variation among realizations of the same strategy. We introduce \textbf{Reward-Conditioned Entropy Allocation (RCEA)}, a hierarchical diagnostic framework that separates complete-trajectory entropy from average token entropy, conditions uncertainty on verifier outcome and process validity, measures correct-solution-mode retention, and connects these population-level quantities to local covariance-driven updates. In controlled sweeps over six RLVR objectives at two model scales, entropy trajectories do not induce a stable ordering of reasoning accuracy. Token entropy is higher for verifier-failed than successful trajectories under both broad separately aggregated estimates and a common-question paired analysis, whose confidence intervals exclude equality at 10 of 12 displayed endpoints. A local covariance approximation predicts the direction of adjacent-checkpoint entropy change in 95.5\% of nonfailed transitions; a reward-stratified endpoint proxy further shows that positive chosen-token covariance is more tail-concentrated within failed prefixes. Every reportable trained endpoint exhibits some contraction in independently audited correct-solution-mode breadth, whose positive association with large-\(k\) coverage remains exploratory. These results show that entropy collapse should be diagnosed by which uncertainty is removed, rather than by global entropy alone.
\end{abstract}

\begin{figure}[!t]
    \centering
    \includegraphics[width=0.94\columnwidth]{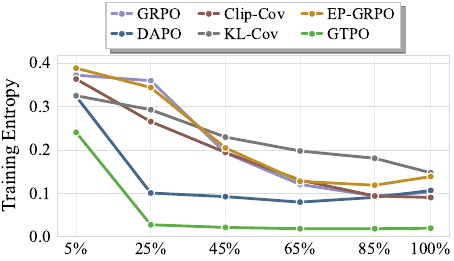}
    \parbox{\columnwidth}{\centering\footnotesize
        \textbf{(a) On-policy token entropy.} Points show centered 50-record
        moving averages at six matched training-progress points.\par}
    \vspace{0.5em}
    \includegraphics[width=0.94\columnwidth]{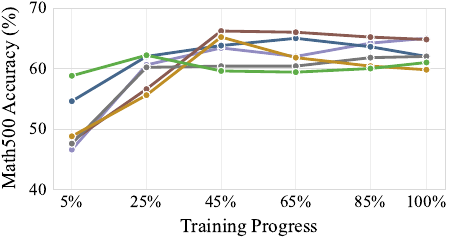}
    \parbox{\columnwidth}{\centering\footnotesize
        \textbf{(b) MATH500 greedy accuracy.} Markers show unsmoothed
        evaluations of 40 problems at the same six progress points.\par}
    \caption{Global entropy does not monotonically track reasoning
        performance in the formal Qwen2.5-3B sweep. }
    \label{fig:teaser}
\end{figure}

\section{Introduction}

Reinforcement learning with verifiable rewards (RLVR) has become a central approach for improving the mathematical and programmatic reasoning of large language models. Given a problem, the policy generates a complete reasoning trajectory, while a rule-based verifier assigns a binary reward according to the final answer \citep{shao2024deepseekmath,deepseekai2025r1}. This sparse but reliable supervision has enabled substantial gains without process annotations. However, when the policy becomes more certain during training, \textbf{has it learned to avoid errors, or has it merely lost the ability to explore alternative correct solutions?}

Recent work suggests that this concern is not merely hypothetical. At the sequence level, \citet{yue2025capacity} find that current RLVR primarily improves sampling efficiency over reasoning paths already accessible to the base model and can reduce large-$k$ problem coverage, narrowing the model's operational reasoning boundary. Average token-level policy entropy is commonly used to diagnose this concentration, but recent analyses reveal that the underlying changes are highly non-uniform. \citet{wang2025forking} identify a small high-entropy minority that acts as critical forks in reasoning trajectories and accounts for most policy-gradient gains, while \citet{cui2025entropy} trace entropy reduction to a small tail of token updates with large positive covariance contributions. Together, these findings suggest that RLVR reallocates probability mass within a pretrained reasoning prior through highly non-uniform local updates. It remains unresolved whether the uncertainty being removed is that of erroneous exploration, redundant realizations, or distinct correct solution strategies.

Figure~\ref{fig:teaser} provides a compact example of the entropy--performance inconsistency. GTPO~\citep{simoni2025gtpo} drives entropy close to zero by 25\% progress while retaining nontrivial MATH500 accuracy. Among the covariance-aware objectives~\citep{cui2025entropy}, Clip-Cov finishes with one of the lowest entropy values and one of the highest accuracies, whereas KL-Cov preserves more late-stage entropy without attaining the best accuracy. Our full two-scale evaluation below yields the same qualitative conclusion: endpoint leaders change with model scale and metric, and their performance ordering does not follow their entropy ordering. Higher global entropy is therefore neither necessary nor sufficient for better RLVR. This challenges the premise that the magnitude of global entropy is itself a reliable measure of useful exploration.

The contradiction arises from conflating distinct quantities within a single scalar. Average token entropy, the statistic commonly logged during training, mixes successful and failed trajectories, reasoning and formatting tokens, and length effects. It is not the entropy of the complete trajectory distribution. Even when restricted to successful trajectories, diversity can reflect either genuinely different mathematical strategies or merely alternative realizations of the same strategy, a distinction formalized by semantic-uncertainty methods \citep{kuhn2023semanticuncertainty,farquhar2024semanticentropy}. Consequently, a decrease in global entropy may signal the elimination of errors, the removal of redundant expression, or the collapse of valuable correct-solution modes; an increase may preserve alternative strategies or merely retain failure-region uncertainty. Global entropy alone cannot distinguish among these cases.

Our central claim is that the magnitude of entropy is not the relevant diagnostic; what matters is how uncertainty is \textbf{allocated} across reward outcomes and semantic solution modes. We formalize this as \textbf{Reward-Conditioned Entropy Allocation (RCEA)}, a three-level diagnostic hierarchy. First, we decompose complete-trajectory and average token entropy by eventual verifier outcome, separating success-region from failure-region uncertainty. Second, within successful trajectories, we group generations by semantic solution strategy, distinguishing diversity across strategies from within-strategy realization variability. Third, we connect these components to local covariance-driven entropy updates, which predict the direction of entropy change while reward conditioning identifies the affected outcome stratum. We evaluate semantic-mode retention as a separate diagnostic and do not estimate mode-stratified covariance. We further introduce a stricter notion of solution-mode collapse, defined through contraction of the conditional distribution over semantically distinct successful modes, and reserve the term \textbf{harmful} collapse for mode contraction that coincides with reduced large-$k$ coverage or out-of-distribution robustness. We evaluate RCEA through controlled comparisons of objectives that induce different entropy trajectories, examining whether independently audited successful-mode retention is exploratorily associated with large-$k$ coverage.

Our contributions are threefold:
\begin{itemize}
	\item We identify and systematically study an entropy--performance inconsistency in RLVR: substantially different global entropy trajectories do not induce a consistent ordering in reasoning accuracy.
	\item We develop RCEA, a hierarchical diagnostic framework that distinguishes complete-trajectory entropy from logged average token entropy, decomposes uncertainty across verifier outcomes and semantic solution modes, and defines solution-mode collapse through contraction of the conditional successful-mode distribution.
	\item We operationalize the framework with fixed-probe estimators, effective mode counts, and reward-stratified covariance attribution, and use them to study greedy performance and, exploratorily, large-$k$ coverage beyond what global entropy reveals.
\end{itemize}

\section{Related Work}
RLVR uses rule-based outcome verification, and GRPO converts sparse outcomes into group-relative advantages without a learned critic \citep{shao2024deepseekmath,deepseekai2025r1,hu2025openreasonerzero,zeng2025simplerlzoo}. DAPO refines this pipeline through dynamic sampling, asymmetric clipping, token-level loss normalization, and overlong-sample control \citep{yu2025dapo}. Because outcome-level acceptance certifies the extracted answer rather than every intermediate step, process-supervision work instead evaluates the reasoning steps themselves \citep{uesato2022solving,lightman2024verify}. Complementing optimizer design, RCEA keeps verifier success and process validity distinct and diagnoses how these objectives redistribute uncertainty.

Large-$k$ evaluations suggest that RLVR mostly sharpens paths accessible to the base policy and may narrow problem coverage \citep{yue2025capacity}. At the token level, a small high-entropy minority marks reasoning forks and accounts for much of the policy-gradient gain \citep{wang2025forking,vassoyan2025criticaltokens,wang2025archer,liu2025hapo}. EP-GRPO gates outcome advantages by predictive entropy and adds an outcome-anchored policy--reference log-probability signal for implicit process guidance \citep{yu2026epgrpo}. Covariance analysis attributes entropy reduction to a small tail of token updates; Clip-Cov and KL-Cov suppress these updates \citep{cui2025entropy}, whereas GTPO controls entropy growth from negative updates to confident trajectories \citep{simoni2025gtpo}. Collectively, these studies connect entropy dynamics to reasoning coverage, influential token updates, and entropy-aware optimization, but do not decompose the affected uncertainty by verifier outcome or determine whether successful trajectories instantiate the same semantic strategy.
Self-consistency benefits from diverse reasoning paths \citep{wang2023selfconsistency}, while semantic-uncertainty methods group meaning-equivalent generations \citep{kuhn2023semanticuncertainty,farquhar2024semanticentropy}. Their primary object is answer-level semantic uncertainty rather than a problem-local partition of the mathematical operations used by successful derivations.

\section{Reward-Conditioned Entropy Allocation}
\label{sec:framework}

\begin{figure*}[t]
    \centering\textit{}
    \includegraphics[width=\textwidth]{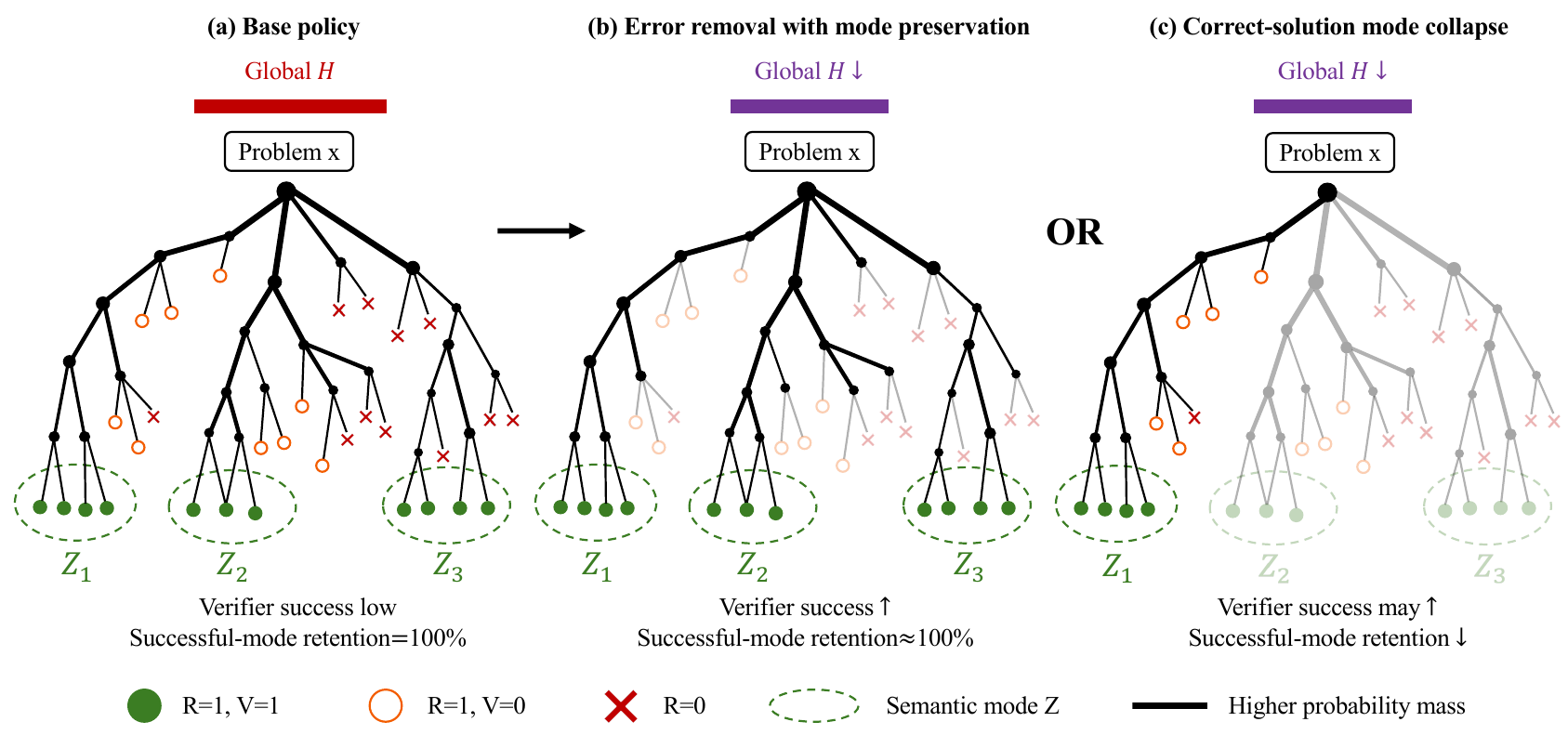}
    \caption{\textbf{Conceptual overview of RCEA.} The same direction of global entropy change can reflect error removal with successful-mode preservation or correct-solution mode contraction. Here \(R\) is verifier reward, \(V\) is process validity, and \(Z\) is a semantic solution mode; the panels illustrate limiting cases rather than empirically assigned collapse labels.}
    \label{fig:rcea-framework}
\end{figure*}

RCEA separates outcome uncertainty, diversity within successful and failed regions, process validity within verifier successes, and diversity among successful semantic strategies. We then connect these global quantities to the local covariance mechanism that changes token entropy. Together, these levels ask where an observed entropy decrease occurs and what kind of behavior it removes. Figure~\ref{fig:rcea-framework} illustrates the resulting distinction. 

\subsection{Preliminary}

Let \(x\) be a reasoning problem and \(Y=(Y_1,\ldots,Y_T)\sim\pi_\theta(\cdot\mid x)\) a complete autoregressive trajectory, including its final answer and end-of-sequence token. An answer extractor \(A_x\) and rule-based verifier \(\mathcal V_x\) define
\begin{equation}
    R=f_x(Y):=\mathcal V_x(A_x(Y))\in\{0,1\}.
    \label{eq:reward}
\end{equation}
Both are deterministic, and the verifier evaluates only the extracted final answer, thus \(H(R\mid Y,x)=0\). Importantly, \(R=1\) certifies verifier acceptance, not the logical validity of every intermediate step. For verifier-successful trajectories, let \(V=u_x(Y)\in\{0,1\}\) denote process validity and \(v_+(x)=P_\theta(V=1\mid R=1,x)\). We model \(V\) as a deterministic target property of the trajectory; experimentally, it is approximated by a frozen judge audited against human labels. We therefore use \textbf{successful} and \textbf{failed} for reward outcomes, reserving \textbf{valid strategy} for modes additionally satisfying \(V=1\).

Let \(p_+(x)=P_\theta(R=1\mid x)\) and \(p_-(x)=1-p_+(x)\). We distinguish two quantities that are often both called policy entropy. The sequence entropy can be written as
\begin{equation}
H_\theta(Y\mid x)=-\sum_y\pi_\theta(y\mid x)\log\pi_\theta(y\mid x).
\end{equation}
It measures uncertainty over complete trajectories. By contrast, training systems usually report a length-normalized average of local next-token entropies.

\subsection{Sequence-Level Reward Decomposition}

\paragraph{Proposition 1 (exact reward decomposition).}
For the deterministic verifier in Eq.~\eqref{eq:reward},
\begin{equation}
\begin{aligned}
H_\theta(Y\mid x)
={}&H_\theta(R\mid x)
+p_+(x)H_\theta(Y\mid R=1,x)\\
&+p_-(x)H_\theta(Y\mid R=0,x).
\end{aligned}
\label{eq:reward-decomposition}
\end{equation}
To see this, expand the joint entropy in two ways:
\begin{equation*}
\begin{aligned}
H(Y,R\mid x)
&=H(Y\mid x)+H(R\mid Y,x)\\
&=H(R\mid x)+H(Y\mid R,x).
\end{aligned}
\end{equation*}
The first conditional term vanishes because \(R=f_x(Y)\), and expanding the second over the two reward values gives Eq.~\eqref{eq:reward-decomposition}.

The first term is binary reward-outcome entropy,
\[
H(R\mid x)=h_{\mathrm b}(p_+)=-p_+\log p_+-p_-\log p_-.
\]
It quantifies uncertainty over success versus failure. It is maximal at \(p_+=1/2\) and approaches zero for both certain success and certain failure; consequently, its decrease is not intrinsically desirable. The remaining terms describe diversity within the successful and failed regions. Equation~\eqref{eq:reward-decomposition} already shows why total entropy is ambiguous: the same decrease can result from resolving reward uncertainty, eliminating failed trajectories, or concentrating the successful distribution.
When \(p_r(x)=0\), the weighted contribution is zero by the usual convention \(0\log 0=0\); reliable estimation of the conditional term requires sufficient observations of outcome \(r\).

\subsection{Reward-Conditioned Token Entropy}

For a sampled trajectory \(y\), define the local entropy at prefix \(s_t=(x,y_{<t})\) and its trajectory-normalized average as
\begin{equation}
    \begin{aligned}
        h_\theta(s_t)&=H\!\left(\pi_\theta(\cdot\mid s_t)\right),\\
        \bar h_\theta(y,x)&=\frac{1}{T(y)}\sum_{t=1}^{T(y)}h_\theta(x,y_{<t}).
    \end{aligned}
    \label{eq:local-token-entropy}
\end{equation}
The on-policy average token entropy is
\(\mathcal H_{\mathrm{tok}}(x)=
\mathbb{E}_{Y\sim\pi_\theta}[\bar h_\theta(Y,x)]\).
By the law of total expectation,
\begin{equation}
    \mathcal H_{\mathrm{tok}}(x)
    =p_+\mathcal H_{\mathrm{tok},+}(x)
    +p_-\mathcal H_{\mathrm{tok},-}(x),
    \label{eq:token-decomposition}
\end{equation}
where
\(\mathcal H_{\mathrm{tok},r}(x)=
\mathbb{E}[\bar h_\theta(Y,x)\mid R=r,x]\).
Unlike Eq.~\eqref{eq:reward-decomposition}, this decomposition has no \(H(R\mid x)\) term, as it partitions an expected local statistic rather than the Shannon entropy of the trajectory distribution.

\subsection{Semantic Solution Modes and Collapse}

Even after process-validity filtering, successful trajectory entropy conflates different mathematical strategies with different realizations of the same strategy. Let \(Z=g_x(Y)\) be a deterministic semantic partition of trajectories satisfying \(R=1,V=1\), where two trajectories share a mode when they use the same core solution strategy. Write \(C_x=(R=1,V=1,x)\). Because \(Z\) is a function of \(Y\), another application of the chain rule gives
\begin{equation}
\begin{aligned}
H(Y\mid C_x)
={}&H(Z\mid C_x)\\
&+\mathbb{E}_{Z\mid C_x}
  [H(Y\mid Z,C_x)].
\end{aligned}
\label{eq:semantic-decomposition}
\end{equation}
The first term is \textbf{valid solution-mode entropy}; the second is \textbf{within-mode realization entropy}. If \(g_x\) is implemented by a stochastic classifier, the right-hand side of Eq.~\eqref{eq:semantic-decomposition} additionally subtracts \(H(Z\mid Y,C_x)\).

Combining Eqs.~\eqref{eq:reward-decomposition} and \eqref{eq:semantic-decomposition} produces the hierarchy
\begin{equation}
\begin{aligned}
H(Y\mid x)
={}&h_{\mathrm b}(p_+)\\
&+p_-H(Y\mid R=0,x)\\
&+p_+\big\{h_{\mathrm b}(v_+)\\
&\quad+v_+H(Z\mid R=1,V=1,x)\\
&\quad+v_+\mathbb{E}_Z[
  H(Y\mid Z,R=1,V=1,x)]\\
&\quad+(1-v_+)H(Y\mid R=1,V=0,x)\big\}.
\end{aligned}
\end{equation}
It separates reward-outcome uncertainty, failed-region diversity, validity uncertainty within verifier successes, valid strategy diversity, within-mode realization diversity, and diversity among verifier-successful but process-invalid trajectories. This is an exact decomposition under the stated deterministic partitions, not an assertion that every component should be maximized.

\paragraph{Definition 1 (correct-solution mode collapse).}
Let
\[
q_{\theta,t}(z\mid x)=P_{\pi_{\theta_t}}(Z=z\mid R=1,V=1,x)
\]
denote the successful mode distribution at training checkpoint \(t\). Its effective number of modes is
\begin{equation}
N^+_{\mathrm{eff},t}(x)
=\exp\!\left(H_{\pi_{\theta_t}}(Z\mid R=1,V=1,x)\right).
\label{eq:effective-modes}
\end{equation}
We say that \textbf{correct-solution mode collapse} occurs when training concentrates \(q_{\theta,t}\) onto a substantially smaller effective set of semantically distinct, valid modes, suppressing modes that had non-negligible probability under the initial policy. Complete collapse is the limiting case \(q_{\theta,t}\to\delta_{z^\star}\), for which \(H(Z\mid R=1,V=1,x)\to0\) and \(N^+_{\mathrm{eff},t}(x)\to1\). A finite decrease is reported as mode contraction and called partial collapse only relative to a predeclared retention threshold. This definition concerns effective probability concentration, not literal support removal, because a softmax policy generally assigns nonzero probability to nominally reachable trajectories.

The definition distinguishes three superficially similar events. Increasing \(p_+\) by moving mass from failed to successful trajectories is successful learning, not collapse. Rejecting process-invalid verifier successes changes \(v_+\), not the conditional mode distribution. Reducing \(H(Y\mid Z,R=1,V=1,x)\) while preserving \(H(Z\mid R=1,V=1,x)\) is realization-level standardization. Only contraction of the conditional semantic distribution removes distinct valid strategies.

We summarize contraction by effective-mode retention relative to the initial policy,
\begin{equation}
\mathrm{Ret}^+_t(x)
=\frac{N^+_{\mathrm{eff},t}(x)}
       {N^+_{\mathrm{eff},0}(x)}.
\label{eq:mode-retention}
\end{equation}

\subsection{From Local Covariance to Entropy Allocation}
The preceding decompositions describe where uncertainty resides, while covariance analysis explains how an optimizer changes local token entropy. For a prefix \(s\), let \(z(v)\) be the logit of token \(v\) and \(\Delta z(v)\) a small update. A first-order expansion gives \citep{cui2025entropy}
\begin{equation}
\Delta h_\theta(s)
\approx-\operatorname{Cov}_{v\sim\pi_\theta(\cdot\mid s)}
\left(\log\pi_\theta(v\mid s),\Delta z(v)\right).
\label{eq:covariance-dynamics}
\end{equation}
Positive covariance therefore decreases local entropy. Under simplified policy-gradient and natural-gradient updates, the corresponding terms take the forms
\(-\eta\operatorname{Cov}(\log\pi,\pi A)\) and
\(-\eta\operatorname{Cov}(\log\pi,A)\), respectively. Large positive contributions are not restricted to high-probability, positive-advantage tokens: low-probability tokens receiving relatively negative updates can also sharpen the distribution. Covariance-aware objectives act on this local mechanism by clipping or regularizing its extreme contributions.

Equation~\eqref{eq:covariance-dynamics}, however, does not determine whether the resulting concentration is useful. It does not identify whether prefix \(s\) belongs to an eventual success or failure, nor whether two successful prefixes instantiate the same semantic strategy. Reward conditioning can group prefix-level entropy and covariance statistics by eventual verifier outcome, while the mode decomposition can in principle separate between-mode from within-mode variation. Our experiments test the reward-stratified covariance bridge and mode retention as separate diagnostics; they do not estimate within-mode realization entropy or mode-stratified covariance. Covariance therefore predicts the \emph{direction} of local entropy change, while conditioning specifies progressively finer questions about its allocation.

This bridge motivates directly testable hypotheses. Lower global entropy can be beneficial when failed-region or within-mode entropy falls while successful mode entropy is retained. Higher global entropy can be unhelpful when it primarily preserves failed-region uncertainty. Harmful collapse is the narrower case in which successful mode entropy and effective mode count contract together, potentially reducing large-\(k\) coverage or out-of-distribution robustness.

\section{Experiments}
\label{sec:experiments}
\begin{table*}[t]
	\centering
	\small
	\setlength{\tabcolsep}{2.8pt}
	\begin{tabular}{lcccccccc}
		\toprule
		Method & AIME'24 & AIME'25 & MATH 500 & AMC'23 & Minerva
		& \shortstack{Average} & \shortstack{Pass@64} & \shortstack{Mode Retention} \\
		\midrule
		\textbf{Qwen2.5-3B}
		& \textbf{13.3} & \textbf{3.3} & 48.6 & 35.0 & 15.0 & 23.05 & 92.8 & \textbf{100.0} \\
		\quad + GRPO \citep{shao2024deepseekmath}
		& 3.3 & \underline{0.0} & 63.8 & 35.0 & \textbf{20.0} & \underline{24.43} & 93.0 & 94.2 \\
		\quad + DAPO \citep{yu2025dapo}
		& \underline{10.0} & \underline{0.0} & \underline{64.4} & 27.5 & \textbf{20.0} & 24.38 & 91.2 & 91.1 \\
		\quad + Clip-Cov \citep{cui2025entropy}
		& \underline{10.0} & \textbf{3.3} & \textbf{65.4} & \textbf{42.5} & \underline{17.5} & \textbf{27.75} & \underline{93.2} & 96.0 \\
		\quad + KL-Cov \citep{cui2025entropy}
		& 3.3 & \textbf{3.3} & 62.6 & 32.5 & 15.0 & 23.35 & \textbf{93.6} & 94.8 \\
		\quad + EP-GRPO \citep{yu2026epgrpo}
		& 0.0 & \underline{0.0} & 61.4 & \underline{37.5} & 15.0 & 22.78 & \underline{93.2} & \underline{99.9} \\
		\quad + GTPO \citep{simoni2025gtpo}
		& 6.7 & \underline{0.0} & 61.0 & \underline{37.5} & 15.0 & 24.03 & 85.6 & 83.7 \\
		\midrule
		\textbf{Qwen2.5-7B}
		& 10.0 & \underline{6.7} & 36.0 & 25.0 & 20.0 & 19.53 & \textbf{94.8} & \textbf{100.0} \\
		\quad + GRPO \citep{shao2024deepseekmath}
		& 10.0 & \underline{6.7} & 71.8 & 45.0 & \textbf{25.0} & 31.69 & 93.0 & 89.6 \\
		\quad + DAPO \citep{yu2025dapo}
		& 13.3 & \underline{6.7} & \textbf{77.2} & 45.0 & \textbf{25.0} & \underline{33.44} & 92.6 & 88.8 \\
		\quad + Clip-Cov \citep{cui2025entropy}\textsuperscript{}
		& 6.7 & 3.3 & 72.8 & 42.5 & 17.5 & 28.56 & -- & 93.5 \\
		\quad + KL-Cov \citep{cui2025entropy}
		& \underline{16.7} & \textbf{13.3} & 68.6 & \textbf{57.5} & \underline{22.5} & \textbf{35.72} & \underline{94.0} & 92.3 \\
		\quad + EP-GRPO \citep{yu2026epgrpo}
		& \textbf{20.0} & \underline{6.7} & \underline{73.2} & 37.5 & 17.5 & 30.97 & \textbf{94.8} & \underline{96.2} \\
		\quad + GTPO \citep{simoni2025gtpo}
		& \underline{16.7} & \underline{6.7} & 71.0 & \underline{47.5} & 15.0 & 31.37 & 89.6 & 85.9 \\
		\bottomrule
	\end{tabular}
	\caption{\textbf{Reasoning accuracy, MATH 500 pass@64, and audited correct-solution-mode retention (\%).} Macro Average weights the five benchmarks equally. Bold and underline mark the best and second-best reportable results within each scale. Clip-Cov-7B: step 3,600 (pre-failure).}
	\label{tab:main-results}
\end{table*}
The RCEA decomposes global entropy into reward-conditioned, semantic-mode, and covariance-attributed components. We now operationalize this hierarchy to ask four questions. 

\subsection{Experimental Setup}
\paragraph{Training Details.}
We train Qwen2.5-3B and Qwen2.5-7B \citep{qwen2024qwen25} for one epoch and one seed on the same 8,000-example subset of Skywork-OR1-RL-Data \citep{skywork2025or1data}. Each prompt receives eight on-policy completions at temperature 1.0 and top-\(p=0.95\); reward is determined solely by the normalized equivalence of the final boxed answer. All objectives share the same LoRA \citep{hu2022lora}, optimizer, and sampling setup within each scale, and training curves use normalized one-epoch progress.

\paragraph{Baseline Methods.}
We compare the untrained base policy with six representative RLVR objectives: GRPO \citep{shao2024deepseekmath}, DAPO \citep{yu2025dapo}, the covariance-aware Clip-Cov and KL-Cov objectives \citep{cui2025entropy}, EP-GRPO \citep{yu2026epgrpo}, and GTPO \citep{simoni2025gtpo}. All non-objective-specific choices are held fixed within each model scale.

\paragraph{Evaluation Strategies.}
We report greedy accuracy on AMC23, AIME24, AIME25, MATH500 \citep{lightman2024verify}, and Minerva Math \citep{lewkowycz2022minerva}, together with MATH500 pass@64 using the standard unbiased estimator \citep{chen2021codex}. A 200-prompt fixed probe sampled at the base policy and four training milestones supports reward-conditioned entropy estimates and correct-solution-mode retention. We report both broad outcome-wise aggregates and a post-hoc common-question comparison restricted to prompts with at least eight samples from each reward outcome. The frozen semantic partition passes an independent adjudicated human audit. We additionally evaluate the exact full-vocabulary covariance prediction across adjacent checkpoints and a reward-stratified endpoint proxy. 

\section{Main Results}
\label{sec:results}
\subsection{RQ1: Does Global Entropy Reliably Indicate Reasoning Performance?}

\begin{figure*}[t]
    \centering
    \includegraphics[width=2\columnwidth]{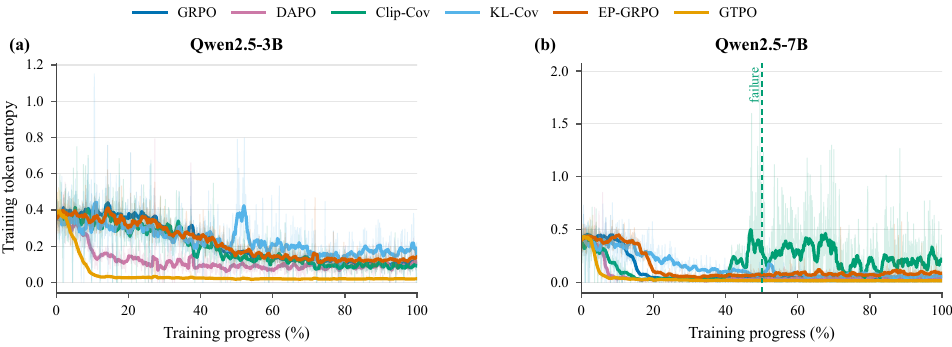}
    \caption{\textbf{Training token-entropy trajectories.} Thin lines are raw traces, thick lines are rolling summaries.}
    \label{fig:training-entropy}
\end{figure*}
Table~\ref{tab:main-results} and Figure~\ref{fig:training-entropy} extend the comparison to benchmark metrics and complete training traces at both scales. Table~\ref{tab:main-results} uses final checkpoints, and no endpoint ranking is stable across scale and metric. Clip-Cov has the highest 3B five-benchmark macro accuracy (27.75\%), whereas KL-Cov leads at 7B (35.72\%). DAPO has the highest 7B MATH500 greedy accuracy (77.2\%), but EP-GRPO and KL-Cov have higher pass@64 (94.8\% and 94.0\% versus 92.6\%). These rankings do not follow the entropy ordering in Figure~\ref{fig:training-entropy}. Most sharply, GTPO drives training entropy close to zero early, yet retains 61.0\% and 71.0\% MATH500 greedy accuracy at 3B and 7B. Conversely, preserving more entropy does not uniformly improve either greedy accuracy or macro accuracy. Global entropy magnitude is therefore an ambiguous diagnostic rather than an objective to maximize.

Notably, Clip-Cov-7B is a distinct training failure: its MATH500 accuracy drops from 60.8\% at step 3,600 to 0.2\% at step 4,000 and remains near zero thereafter. This transition coincides with completion lengths saturating the token limit and the boxed-answer rate collapsing from 95.2\% to 1.0\%. During the surrounding failure window, most updates activate neither reward-variance learning signals nor the covariance mask, suggesting a self-reinforcing near-all-failure regime in which the sparse Clip-Cov intervention provides little recovery signal. 

\subsection{RQ2: How Is Entropy Allocated Across Successful and Failed Trajectories?}

\begin{figure}[t]
    \centering
    \includegraphics[width=\columnwidth]{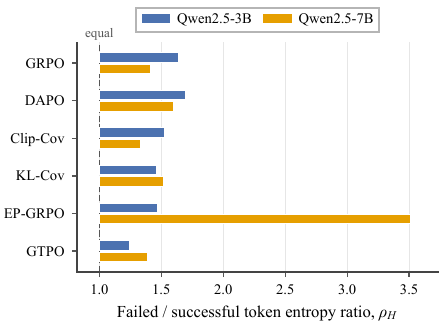}
    \caption{\textbf{Failed-to-successful token-entropy ratios.} Values above one indicate higher entropy in verifier-failed trajectories; outcome strata are aggregated separately.}
    \label{fig:reward-entropy}
\end{figure}

Figure~\ref{fig:reward-entropy} summarizes the broad reward-stratified ordering through \(\rho_H=\mathcal{H}_{\mathrm{tok},-}/\mathcal{H}_{\mathrm{tok},+}\). All 12 displayed endpoints have \(\rho_H>1\). However, successful and failed coverage ranges from 40.5--53.0\% and 72.5--85.5\%, respectively, so the two aggregates need not use the same problems. Near-zero reconstruction residuals verify the decomposition's accounting but not this substantive ordering.

\begin{figure}[t]
    \centering
    \includegraphics[width=\columnwidth]{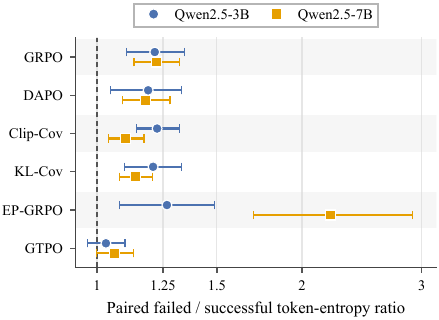}
    \caption{\textbf{Common-question paired token-entropy ratios.} Points show geometric means and bars denote paired-bootstrap 95\% confidence intervals.}
    \label{fig:reward-entropy-paired}
\end{figure}

To control for prompt-set composition, Figure~\ref{fig:reward-entropy-paired} compares both outcomes within the same 40--57 eligible questions per endpoint. All 12 paired estimates exceed one, and bootstrap intervals exclude equality for 10; only GTPO-3B and GTPO-7B overlap one. The broad ordering therefore persists after controlling for question identity, although the paired result does not cover prompts dominated by a single outcome. Together, the analyses show that verifier-failed trajectories generally retain more token-level uncertainty than successful trajectories, motivating future investigation of process-aware rewards.

\subsection{RQ3: Is Correct-Solution-Mode Retention Associated with Large-$k$ Coverage?}

Table~\ref{tab:main-results} shows that audited effective-mode retention ranges from 83.7\% to 99.9\% at 3B and from 85.9\% to 96.2\% at 7B. All algorithms exhibited a reduction in the patterns of correct solutions. This indicates that, at the current stage, RLVR algorithms narrow the scope of reasoning.  Across trained methods with reportable coverage, its association with MATH500 pass@64 is positive (Spearman \(\rho=0.812\), \(n=6\), at 3B; \(\rho=1.000\), \(n=5\), at 7B). EP-GRPO has the highest retention at both scales but not the highest greedy accuracy, while DAPO-7B retains fewer modes than GRPO-7B yet has substantially higher greedy MATH500 accuracy (77.2\% versus 71.8\%). Thus, the association is consistent with strategy breadth benefiting repeated-sampling coverage.

\subsection{RQ4: Does Local Covariance Explain Entropy Change and Its Reward-Stratified Concentration?}

\begin{figure}[t]
    \centering
\includegraphics[width=0.7\columnwidth]{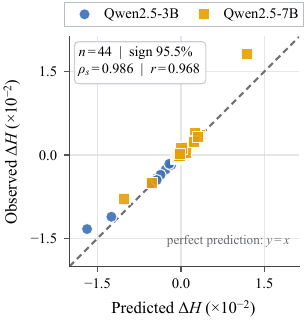}
    \caption{Predicted versus observed entropy changes across adjacent checkpoints. \(y=x\) indicates perfect prediction.}
    \label{fig:covariance-fit}
\end{figure}

\paragraph{Predicting entropy change.}
Across the 44 adjacent-checkpoint transitions from 11 nonfailed runs, the covariance approximation predicts the sign of the observed entropy change in 42 cases (95.5\%), with Spearman \(\rho=0.986\), Pearson \(r=0.968\), slope \(0.993\), and \(R^2=0.937\) (Figure~\ref{fig:covariance-fit}). The median relative error is 12.7\%. The points closely follow the identity line \(y=x\), indicating that the approximation captures not only the direction but also the magnitude of the observed entropy change. This near-identity fit provides strong consistency evidence for the local mechanism in Eq.~\eqref{eq:covariance-dynamics}.

\paragraph{Locating reward-stratified tail concentration.}
Table~\ref{tab:covariance-tail} shows the same separation at both scales and for every objective. Across trained, nonfailed methods, the q99 tail accounts for 4.6--18.2\% of positive chosen-token covariance within successful-prefix strata but 70.4--99.3\% within failed-prefix strata, gaps of 65.8--83.7 percentage points. Because cutoffs are estimated separately by reward stratum, the values compare concentration rather than absolute covariance mass. Ties are retained; for GTPO failed-prefix strata they select roughly 58--59\% of tokens, so this is not a strict top-1\% attribution. Figure~\ref{fig:covariance-fit} tests \emph{how} covariance predicts entropy change, whereas Table~\ref{tab:covariance-tail} identifies where its positive endpoint contribution is most tail-concentrated. Together they support the local-to-reward-stratified diagnostic bridge.

\begin{table}[t]
    \centering
    \small
    \setlength{\tabcolsep}{3.2pt}
    \renewcommand{\arraystretch}{1.08}
    \begin{tabular*}{\columnwidth}{@{\extracolsep{\fill}}lcc@{}}
        \toprule
        Method &
        \shortstack{Qwen2.5-3B\\[-1pt]\scriptsize successful \(\rightarrow\) failed} &
        \shortstack{Qwen2.5-7B\\[-1pt]\scriptsize successful \(\rightarrow\) failed} \\
        \midrule
        Base     & \shortstack{2.4 \(\rightarrow\) 61.2\\[-1pt]\scriptsize \(+58.8\) pp} & \shortstack{2.7 \(\rightarrow\) 43.2\\[-1pt]\scriptsize \(+40.4\) pp} \\
        GRPO     & \shortstack{9.5 \(\rightarrow\) 82.7\\[-1pt]\scriptsize \(+73.2\) pp} & \shortstack{9.9 \(\rightarrow\) 91.8\\[-1pt]\scriptsize \(+81.9\) pp} \\
        DAPO     & \shortstack{12.5 \(\rightarrow\) 83.9\\[-1pt]\scriptsize \(+71.4\) pp} & \shortstack{10.8 \(\rightarrow\) 93.2\\[-1pt]\scriptsize \(+82.4\) pp} \\
        Clip-Cov & \shortstack{10.0 \(\rightarrow\) 84.5\\[-1pt]\scriptsize \(+74.5\) pp} & \shortstack{1.5 \(\rightarrow\) 83.0\({}\)\\[-1pt]\scriptsize \(+81.4\) pp} \\
        KL-Cov   & \shortstack{4.6 \(\rightarrow\) 70.4\\[-1pt]\scriptsize \(+65.8\) pp} & \shortstack{10.2 \(\rightarrow\) 86.2\\[-1pt]\scriptsize \(+76.0\) pp} \\
        EP-GRPO  & \shortstack{7.1 \(\rightarrow\) 82.9\\[-1pt]\scriptsize \(+75.8\) pp} & \shortstack{7.6 \(\rightarrow\) 81.3\\[-1pt]\scriptsize \(+73.6\) pp} \\
        GTPO     & \shortstack{18.2 \(\rightarrow\) 98.6\\[-1pt]\scriptsize \(+80.4\) pp} & \shortstack{15.6 \(\rightarrow\) 99.3\\[-1pt]\scriptsize \(+83.7\) pp} \\
        \bottomrule
    \end{tabular*}
    \caption{\textbf{Within-stratum q99 share of positive chosen-token covariance (\%), successful \(\rightarrow\) failed.}}
    \label{tab:covariance-tail}
\end{table}

\raggedbottom
\section{Discussion}

Together, the four RQs explain why global entropy is an ambiguous RLVR diagnostic. RQ1 shows that it does not stably order reasoning performance, while RQ2 localizes part of this ambiguity to different uncertainty levels across verifier outcomes. RQ4 links the observed changes to local covariance and its concentration in failed prefixes; RQ3 asks the distinct semantic question of whether correct strategies are retained. Covariance thus explains how a distribution sharpens locally, reward conditioning identifies where this occurs, and mode analysis asks what behavior is removed.

Entropy should be neither universally maximized nor minimized. A decrease may remove failed-region uncertainty or redundant realizations, but becomes harmful when it eliminates distinct correct strategies. The Clip-Cov-7B failure likewise shows why entropy must be interpreted alongside reward variance, response length, and termination behavior. Although diagnostic rather than causal, this pattern motivates adaptive controls based on outcome composition instead of a single global entropy target. RCEA accordingly pairs global entropy with reward-conditioned summaries, length diagnostics, audited mode retention, covariance attribution.

\section{Conclusion and Limitations}

Global token entropy conflates uncertainty across reward outcomes and solution strategies, and therefore does not consistently order RLVR reasoning performance. We introduce RCEA to resolve this ambiguity through trajectory-level decomposition, reward conditioning, independently audited solution modes, and local covariance attribution. Across six objectives and two model scales, failed trajectories retain higher token entropy, covariance closely predicts local entropy change and is most tail-concentrated in failed prefixes, and all reportable trained endpoints show estimated contraction in correct-solution-mode breadth. The exploratory association between mode retention and pass@64 further suggests that strategy breadth may matter more for repeated-sampling coverage than for greedy accuracy. RCEA thus replaces a one-dimensional collapse signal with a structured diagnosis of where uncertainty changes and whether correct reasoning diversity is preserved.

The study uses a single seed, one model family at two scales. Process validity remains model-judged, the paired and mode analyses cover restricted prompt sets, and we do not estimate within-mode realization entropy or mode-stratified covariance. The finite-difference covariance analysis is diagnostic, leaving causal validation to future work.

\bibliography{aaai2027}


\end{document}


\maketitle

\section{Scope and Reporting Conventions}

This technical supplement supplies measurement definitions, implementation details, audit
information, and extended endpoint summaries for the main paper.  It does not
introduce additional claims.  We use \emph{successful} and \emph{failed} for
the binary verifier outcome \(R\), and \emph{process-valid} only for a
verifier-successful trajectory accepted by the frozen validity judge.  All
reported entropies use natural logarithms.  Raw training step denotes prompt
exposure: one 3B rollout contains eight prompt groups and one 7B rollout
contains four, so one epoch ends at raw steps 4,000 and 8,000, respectively.
Training-progress percentages are used whenever scales are compared.

\section{Measurement and Estimation}
\label{sec:sup-measurement}

\subsection{Hierarchical Quantity Being Measured}

Because \(R=f_x(Y)\), \(H(R\mid Y,x)=0\).  With
\(p_+(x)=P(R=1\mid x)\), \(p_-=1-p_+\), and
\(v_+(x)=P(V=1\mid R=1,x)\), repeated application of the chain rule gives
\begin{equation}
\begin{aligned}
H(Y\mid x)
={}&h_{\mathrm b}(p_+)+p_-H(Y\mid R=0,x)\\
&+p_+\big\{h_{\mathrm b}(v_+)
+v_+H(Z\mid R=1,V=1,x)\\
&\quad+v_+\mathbb E_Z[H(Y\mid Z,R=1,V=1,x)]\\
&\quad+(1-v_+)H(Y\mid R=1,V=0,x)\big\}.
\end{aligned}
\label{eq:sup-hierarchical-decomposition}
\end{equation}
The identity separates reward uncertainty, failed-region diversity, validity
uncertainty, between-mode diversity, within-mode realization diversity, and
diversity among verifier-successful but process-invalid trajectories.  It is
an accounting identity, not a claim that every component should be maximized.
The experiments estimate reward-conditioned token and sequence statistics and
successful-mode retention.  They do not estimate within-mode realization
entropy or mode-stratified covariance.

\subsection{Fixed-Probe Sampling}

Let \(\mathcal X\) be the fixed 200-question diagnostic set.  At the base
policy and at 25\%, 50\%, 75\%, and 100\% training progress, we draw
\(K=64\) independent completions per \(x\in\mathcal X\) using temperature
1.0 and top-\(p=1\).  The same questions, decoding rule, 2,048-token
generation limit, and verifier are used at every checkpoint.  For completion
\(i\), we store its verifier reward \(r_i\), completion length \(T_i\),
truncation indicator, trajectory log probability
\[
\ell_i=\log\pi_{\theta_t}(y_i\mid x)
      =\sum_{j=1}^{T_i}\log\pi_{\theta_t}
       (y_{i,j}\mid x,y_{i,<j}),
\]
and trajectory-normalized token entropy
\[
\bar h_i=\frac{1}{T_i}\sum_{j=1}^{T_i}
H\!\left(\pi_{\theta_t}(\cdot\mid x,y_{i,<j})\right).
\]
The end-of-sequence token is included when it is generated.  Sampled
trajectories are replayed with teacher forcing so that token entropies and log
probabilities are evaluated by the checkpoint that generated them.  We retain
on-policy entropy logged during training as an optimization trace, but use the
fixed probe for cross-objective conditional comparisons.

Metrics are first computed within question and then macro-averaged.  This
prevents an easy question with many successful samples from dominating an
outcome-conditioned estimate.  For
\(K_r(x)=\sum_i\mathbf 1(r_i=r)\), a conditional quantity is reported only
when \(K_r(x)\geq8\); its prompt coverage is always reported.  Rare outcomes
are omitted rather than assigned zero entropy.

\subsection{Reward-Conditioned Estimators}

The empirical reward probability and outcome entropy are
\[
\widehat p_r(x)=\frac{K_r(x)}{K},
\qquad
\widehat H_R(x)=h_{\mathrm b}(\widehat p_+(x)).
\]
The principal token statistic gives each trajectory equal weight:
\begin{equation}
\widehat{\mathcal H}_{\mathrm{tok},r}(x)
=\frac{1}{K_r(x)}\sum_{i:r_i=r}\bar h_i.
\label{eq:sup-token-estimator}
\end{equation}
We additionally retain the pooled-token average, mean response length, and
truncation rate to identify length-induced discrepancies.  The
sequence-level conditional entropy is estimated from the same on-policy
samples.  Since
\(\log\pi_\theta(y\mid R=r,x)=
\log\pi_\theta(y\mid x)-\log p_r(x)\),
\begin{equation}
\widehat H_{\mathrm{seq},r}(x)
=-\frac{1}{K_r(x)}\sum_{i:r_i=r}\ell_i
+\log\widehat p_r(x).
\label{eq:sup-sequence-estimator}
\end{equation}
This estimator preserves the sequence-level meaning of the chain-rule
decomposition but is length-sensitive and higher variance.  We therefore use
it primarily as an accounting check and center the cross-objective analysis
on Eq.~\eqref{eq:sup-token-estimator}.  Components constructed from the same
samples reconstruct \(-K^{-1}\sum_i\ell_i\) and the unconditional token
average up to floating-point error; the observed mean absolute residuals are
\(2.74\times10^{-15}\) and \(7.63\times10^{-19}\), respectively.

\subsection{Broad and Common-Question Comparisons}

The broad ratio in the main paper is
\[
\rho_H=
\frac{\operatorname{MacroAvg}_{x:K_-(x)\geq8}
      \widehat{\mathcal H}_{\mathrm{tok},-}(x)}
     {\operatorname{MacroAvg}_{x:K_+(x)\geq8}
      \widehat{\mathcal H}_{\mathrm{tok},+}(x)}.
\]
Its numerator and denominator may cover different questions, which increases
coverage but makes the comparison descriptive.  The paired analysis retains
only questions satisfying \(K_+(x),K_-(x)\geq8\), computes
\(d_x=\log\widehat{\mathcal H}_{\mathrm{tok},-}(x)-
\log\widehat{\mathcal H}_{\mathrm{tok},+}(x)\), and reports
\(\exp(|\mathcal X_{\mathrm{pair}}|^{-1}\sum_x d_x)\).  Its 95\% interval
uses 10,000 paired question-bootstrap replicates.  This was a post-hoc
robustness analysis and does not cover questions dominated by one reward
outcome.

\subsection{Semantic-Mode Estimation}

For judge-accepted \(R=1,V=1\) trajectories, let \(n_{t,z}(x)\) be the count
assigned to problem-local mode \(z\), with
\(K_{+,V}(x)=\sum_z n_{t,z}(x)\).  We estimate
\begin{equation}
\begin{aligned}
\widehat q_t(z\mid x)&=\frac{n_{t,z}(x)}{K_{+,V}(x)},\\
\widehat H^+_{Z,t}(x)
&=-\sum_z\widehat q_t(z\mid x)\log\widehat q_t(z\mid x),\\
\widehat N^+_{\mathrm{eff},t}(x)
&=\exp\!\left(\widehat H^+_{Z,t}(x)\right).
\end{aligned}
\label{eq:sup-mode-estimator}
\end{equation}
Mode retention is
\begin{equation}
\widehat{\mathrm{Ret}}^+_t(x)
=\frac{\widehat N^+_{\mathrm{eff},t}(x)}
       {\widehat N^+_{\mathrm{eff},0}(x)}.
\label{eq:sup-mode-retention}
\end{equation}
Because plug-in entropy is downward biased and the number of accepted
successes differs across checkpoints, each comparison is rarefied to a common
accepted-success count before taking the ratio.  We then macro-average over
reportable questions.  The set of reportable questions is method-dependent,
so the resulting method-level association with MATH500 pass@64 is
exploratory.  Retention is treated continuously.  A formal partial-collapse
label would additionally require a predeclared retention threshold and
suppression of a mode that was non-negligible under the base policy; the
present paper does not instantiate that stronger label.

\subsection{Covariance Attribution}

The exact analysis evaluates the full vocabulary at identical prefixes under
an old checkpoint and the checkpoint 50 raw steps later.  It uses 40 fixed
questions, four sampled trajectories per question, and up to 64 eligible token
positions per trajectory.  For each transition, full-vocabulary covariance is
aggregated over sampled positions and compared with the directly observed
finite entropy difference.  These 50-step differences test a local
approximation over a short finite interval rather than an infinitesimal
optimizer update.

The reward-stratified analysis is separate.  At a selected endpoint it
reconstructs the common binary group advantage and computes a chosen-token
covariance proxy for every eligible prefix.  Within each reward stratum, q99
means the set of values at or above that stratum's empirical 99th-percentile
cutoff.  We report the fraction of positive covariance mass contained in this
set.  Cutoffs are fitted separately for success and failure, so the statistic
compares concentration rather than absolute covariance mass.  Ties are
retained: for GTPO failed-prefix strata, the q99 set contains approximately
58--59\% of token positions and is therefore not literally a top-1\% subset.
The endpoint proxy neither apportions total entropy change between reward
strata nor establishes causality.

\subsection{Uncertainty and Failure Exclusion}

Where intervals are reported, we use hierarchical resampling at the available
question and trajectory levels.  With one training seed, these intervals
reflect prompt and sampling uncertainty, not seed-to-seed variability.
Smoothing parameters for training curves are fixed for visualization and are
not selected using downstream accuracy.  The failed Clip-Cov-7B run is
excluded from cross-method correlations and from the exact covariance fit,
but its final endpoint is retained as a marked diagnostic case in the q99
table.

\section{Experimental Configuration}

\subsection{Training Data and Shared Optimization}

The 8,000 training examples are sampled once from Skywork-OR1-RL-Data
\citep{skywork2025or1data} with dataset seed 42, after removing records
without an extractable answer and prompts exceeding 10\% of the 2,048-token
context.  The policy is instructed to reason step by step and place its final
answer in \verb|\boxed{}|.  The binary reward is determined only by normalized
equivalence of the extracted final answer.

Training uses TRL \citep{vonwerra2020trl} and Transformers
\citep{wolf2020transformers}, with vLLM \citep{kwon2023vllm} for colocated
rollout generation.  All runs use seed 42, bfloat16 weights, gradient
checkpointing, and LoRA \citep{hu2022lora} rank 32 with scale 64 on every
attention and MLP projection.  Optimization uses 8-bit AdamW
\citep{loshchilov2019adamw,dettmers2022eightbit}, learning rate
\(5\times10^{-6}\), weight decay \(10^{-3}\), and 10\% linear warmup.  Four
gradient-accumulation steps yield batches of 64 trajectories for 3B and 32 for
7B.  We use four optimizer iterations per rollout, token-level importance
ratios, no standard reference-policy KL penalty, and checkpoints every 50 raw
steps.  Rollouts use eight completions per prompt at temperature 1.0 and
top-\(p=0.95\).

\subsection{Objective-Specific Settings}

GRPO uses symmetric clipping at 0.2.  DAPO uses token-level loss
normalization, asymmetric clipping \([0.2,0.28]\), dynamic removal of
zero-variance groups, and soft overlong shaping over the final 20\% of the
completion budget.  Clip-Cov selects the covariance tail at ratio
\(2\times10^{-4}\), with lower and upper factors 1 and 5.  KL-Cov uses
selection ratio \(2\times10^{-3}\) and coefficient 0.5.  EP-GRPO uses 10
relative-position buckets, entropy-gate coefficient 3.0, progress weight 0.2,
and process scale 1.0.  GTPO uses entropy coefficient 0.1 and threshold
\(\log2\), with the initial completion entropy calibrated from the first
on-policy batch.

\subsection{Compute Infrastructure and Software Environments}

\begin{table*}[t]
\centering
\scriptsize
\setlength{\tabcolsep}{3pt}
\begin{tabular}{@{}p{0.10\textwidth}p{0.20\textwidth}p{0.24\textwidth}p{0.11\textwidth}p{0.24\textwidth}@{}}
\toprule
Role & GPU & CPU & System memory & Execution pattern\\
\midrule
Training &
1\(\times\) RTX PRO 6000 (96 GB) &
25 vCPU, Xeon Platinum 8470Q &
120 GB &
One single-GPU training job\\
Inference &
8\(\times\) RTX 5090 D (32 GB each) &
200 vCPU, Xeon Platinum 8470Q &
496 GB &
Up to eight independent workers\\
\bottomrule
\end{tabular}
\caption{Compute infrastructure used for the reported experiments.  The
eight inference GPUs execute independent single-GPU tasks; Qwen2.5-3B/7B do
not use tensor parallelism.}
\label{tab:sup-compute}
\end{table*}

The training server executes the twelve RLVR runs, whereas the inference
server performs trajectory generation, teacher-forced entropy measurement,
covariance diagnostics, and the local Qwen3.5 validity judge.  The
Qwen3.7-Flash codebook stage uses a remote OpenAI-compatible API and is
API-bound rather than GPU-bound.  Aggregation and frozen-data reproduction
are CPU-only.

Both experimental environments use the supplied Python 3.11.0 runtime record
and Linux containers.  The training environment contains PyTorch 2.8.0,
Transformers 4.57.6, TRL 0.29.0, vLLM 0.11.0, PEFT 0.18.1, and Datasets
4.6.1.  The inference environment contains PyTorch 2.11.0, Transformers
5.12.1, vLLM 0.23.0, PEFT 0.19.1, NumPy 2.3.5, and scikit-learn 1.9.0.
Unmodified \texttt{pip freeze} snapshots for both servers are included in
the Code and Data Supplement. 

\subsection{Run Counts, Randomness, and Parameter Selection}

Each method--scale entry is one training run: six objectives at two scales
give 12 trained runs, all with training seed 42.  Base-policy rows require no
training.  Dataset filtering and the 8,000-example draw use seed 42.  The
fixed-probe sample uses seed 2027, and MATH500 pass@\(k\) uses inference
seeds 0--63.  The validity judge uses temperature zero and seed 2027.
Codebook ordering uses fixed seeds 2027, 2028, and 2029 for induction, merge,
and final assignment; rarefaction and independent-audit sampling use seeds
2030 and 2031.  API responses at temperature zero can still vary with the
hosted implementation, so assignments are repeated twice and disagreements
are reconciled rather than assumed deterministic.

We did not conduct a systematic hyperparameter sweep for the reported
comparison.  Shared optimization settings and published method defaults were
fixed before formal benchmark evaluation; the final objective-specific
values are listed above and are serialized in
\texttt{configs/experiment.json}.  No test-set result is used to select among
multiple training seeds.  The Code and Data Supplement contains preprocessing,
training, measurement, aggregation, and plotting code plus a source-to-paper
map.  Public datasets, model weights, checkpoints, external API service, and
human annotation labor are not redistributed or automated.

\subsection{Benchmark and Probe Inventory}

\begin{table}[t]
\centering
\small
\setlength{\tabcolsep}{3.1pt}
\begin{tabular*}{\columnwidth}{@{\extracolsep{\fill}}lrrl@{}}
\toprule
Dataset & Benchmark & Probe & Main use\\
\midrule
AMC23 & 40 & 20 & Greedy, probe\\
AIME24 & 30 & 20 & Greedy, probe\\
AIME25 & 30 & 20 & Greedy, probe\\
MATH500 & 500 & 100 & Greedy, pass@\(k\), probe\\
Minerva Math & 40 subset & 40 & Greedy, probe\\
\bottomrule
\end{tabular*}
\caption{Evaluation inventory.  ``Probe'' is the number of fixed questions
used for entropy and semantic-mode diagnostics.}
\label{tab:sup-inventory}
\end{table}

Single-sample benchmark accuracy uses greedy decoding with a 2,048-token
completion limit.  MATH500 is additionally evaluated at 20 equally spaced
progress points; all five benchmarks are evaluated at 25\%, 50\%, 75\%, and
100\% progress.  Final pass@\(k\), for
\(k\in\{1,4,8,16,32,64\}\), uses 64 independently seeded samples per MATH500
question at temperature 0.7 and top-\(p=0.95\), with the standard unbiased
estimator \citep{chen2021codex}.  The ``Macro Average'' in the main table is
the unweighted mean of the five benchmark accuracies; it does not weight
datasets by their number of questions.

\subsection{Checkpoint Selection and the Clip-Cov-7B Exception}

\begin{table*}[t]
\centering
\small
\setlength{\tabcolsep}{4pt}
\begin{tabular*}{\textwidth}{@{\extracolsep{\fill}}lll}
\toprule
Main-paper artifact & Ordinary reporting point & Clip-Cov-7B treatment\\
\midrule
Training-entropy trace & Complete one-epoch trace & Complete trace; failure marked at step 4,000\\
Main benchmark table & Final checkpoint & Step 2,000 pre-collapse milestone\\
Broad/paired reward ratios & Final diagnostic checkpoint & Step 2,000 pre-collapse probe\\
Exact covariance fit & Four adjacent 50-step transitions & All four transitions excluded\\
Reward-stratified q99 table & Final endpoint proxy & Step 8,000 retained as failed diagnostic\\
\bottomrule
\end{tabular*}
\caption{Checkpoint policy.  Different Clip-Cov-7B entries answer different
questions and must not be interpreted as measurements of one common endpoint.}
\label{tab:sup-checkpoints}
\end{table*}

For every nonfailed trained run, endpoint benchmark and probe results use the
final checkpoint: raw step 4,000 for 3B and 8,000 for 7B.  Clip-Cov-7B is the
only exception.  Its five greedy entries and Mode Retention in the main
benchmark table use step 2,000, the last common five-benchmark milestone before
the detected collapse; later performance checkpoints are excluded from the
comparative ranking.  Pass@64 was not run at step 2,000 and is therefore
unreported.  The reward-conditioned figures likewise use its step-2,000
probe.  Conversely, the q99 table deliberately retains the step-8,000
chosen-token endpoint proxy as a failure diagnostic.  This explains why its
q99 row is not part of the numerical range quoted for trained, nonfailed
methods.

\section{Validity Filtering and Semantic Codebooks}

\subsection{Process-Validity Judge}

Verifier acceptance establishes only that the extracted final answer matches
the reference.  A frozen Qwen3.5-9B judge
\citep{alibaba2026modelstudio} therefore evaluates whether a
verifier-successful trajectory contains a fatal reasoning error.  Calibration
uses 600 blinded trajectories balanced for valid and invalid examples.  On a
117-item held-out split, the judge obtains 95.7\% accuracy and macro-F1,
94.9\% accepted-valid precision, 96.6\% valid recall, and complete coverage.
An independent accepted-set check estimates accepted-valid precision at
91.3\% and 89.1\% against two annotators.  These checks estimate precision
within automatically accepted trajectories, not recall over all invalid
trajectories.  We consequently use \emph{judge-accepted process-valid} rather
than \emph{ground-truth valid}.

Across all fixed-probe checkpoints and objectives, the rule-based verifier
accepts 118,435 trajectories.  The judge retains 92,867 trajectories from 125
questions.  Records are joined by a stable trajectory identifier, and
provenance fields identifying method, model scale, checkpoint, and decoding
seed are hidden from all codebook-induction prompts.

\subsection{Question-Local Codebook Pipeline}

Semantic modes are defined separately for each problem; trajectories from
different problems are never clustered together.  Two trajectories share a
mode when their decisive mathematical method and core sequence of operations
are the same.  Differences in wording, notation, arithmetic detail, or order
of exposition do not create a new mode unless they change the underlying
solution method.

Qwen3.7-Flash \citep{alibaba2026modelstudio} first induces local candidate
modes from 3,941 blinded trajectory chunks.  The resulting 6,586 candidate
definitions are merged question by question and frozen before final
assignment.  The merge yields 125 codebooks and 307 initial modes (one to
eight modes per question; median two).  Every accepted trajectory is then
assigned twice against the frozen codebook in 15,604 batches.  A trajectory is
sent to reconciliation when the local assignments disagree, the two final
passes disagree, a special label is returned, or confidence is low.  This
produces 7,726 disputed trajectories and 1,315 arbitration batches.  Nine
provisional ``NEW'' assignments are reviewed in five question-level extension
requests.  Confirmed extensions are added and the affected codebook is
refrozen.  The final export contains 313 modes, 92,861 included trajectories,
and six uncertain trajectories.

The pipeline is intentionally not an embedding-threshold clustering method.
It constructs explicit problem-local definitions, assigns against a frozen
codebook, and reconciles disagreements.  This avoids treating one global
cosine threshold as a ground-truth semantic boundary and prevents
uncontrolled codebook drift from appearing as mode creation or disappearance.

\subsection{Independent Partition Audit}

The independent audit is question-disjoint from development labels and
contains 29 questions and 4,526 within-question trajectory pairs.  Two
annotators independently mark each pair as using the same or a different core
strategy.  Before adjudication, human--human macro-F1 is 0.859; automatic
agreement is 0.777 against annotator A and 0.894 against annotator B.  The
preliminary discrepancy is driven primarily by different granularity in
splitting same-strategy variants.

The frozen rubric predeclares five requirements: human--human macro-F1 at
least 0.85; automatic macro-F1 at least 0.85; an automatic-to-human gap no
larger than 0.05; precision and recall for the ``different'' class each at
least 0.80; and automatic trajectory coverage at least 0.95.  Disagreements
are adjudicated under the frozen rubric rather than by changing model
assignments after observing an endpoint result.  The final adjudicated
comparison satisfies the predeclared gate and is the basis for the
``audited'' terminology in the main paper.  The preliminary values are
reported above because they show the annotator sensitivity that necessitated
adjudication.  The audit validates a sampled partition; it does not amount to
manual verification of all 92,861 included trajectories.

\section{Extended Endpoint Diagnostics}

\subsection{Reward-Conditioned Entropy}

\begin{table*}[t]
\centering
\scriptsize
\setlength{\tabcolsep}{3.2pt}
\begin{tabular*}{\textwidth}{@{\extracolsep{\fill}}llrrrr}
\toprule
Scale & Method & Broad \(\rho_H\) & Success/failure coverage &
\(n_{\mathrm{paired}}\) & Paired ratio [95\% CI]\\
\midrule
3B & GRPO     & 1.642 & 43.5/82.5 & 52 & 1.216 [1.105, 1.345]\\
3B & DAPO     & 1.699 & 42.0/80.5 & 45 & 1.189 [1.048, 1.332]\\
3B & Clip-Cov & 1.526 & 44.5/83.5 & 56 & 1.225 [1.143, 1.321]\\
3B & KL-Cov   & 1.457 & 43.5/85.0 & 57 & 1.209 [1.098, 1.329]\\
3B & EP-GRPO  & 1.467 & 40.5/85.5 & 52 & 1.267 [1.079, 1.487]\\
3B & GTPO     & 1.244 & 41.5/79.5 & 42 & 1.031 [0.969, 1.099]\\
\midrule
7B & GRPO     & 1.416 & 50.5/75.5 & 52 & 1.223 [1.133, 1.323]\\
7B & DAPO     & 1.598 & 49.0/73.0 & 44 & 1.179 [1.091, 1.280]\\
7B & Clip-Cov & 1.329 & 52.0/73.5 & 51 & 1.102 [1.039, 1.172]\\
7B & KL-Cov   & 1.517 & 53.0/75.0 & 56 & 1.138 [1.079, 1.208]\\
7B & EP-GRPO  & 3.512 & 49.5/79.0 & 57 & 2.204 [1.696, 2.909]\\
7B & GTPO     & 1.391 & 47.5/72.5 & 40 & 1.061 [0.998, 1.130]\\
\bottomrule
\end{tabular*}
\caption{Endpoint failed-to-successful token-entropy ratios.  Coverage is the
percentage of the 200 probe questions satisfying \(K_r\geq8\).  Clip-Cov-7B
uses the pre-collapse step-2,000 probe; all other trained runs use their final
probe.}
\label{tab:sup-reward-entropy}
\end{table*}

The broad estimate is above one at all 12 displayed endpoints, but its
outcome-specific coverage sets differ.  The paired estimate controls question
identity and is also above one at all endpoints; the intervals for GTPO at
both scales overlap equality.  These are endpoint associations.  They do not
show that high failed-region entropy causes failure, nor that reducing it
would necessarily improve accuracy.

\begin{table}[t]
\centering
\scriptsize
\setlength{\tabcolsep}{3pt}
\begin{tabular*}{\columnwidth}{@{\extracolsep{\fill}}llrrr@{}}
\toprule
Scale & Method & \(n\) & Failure higher & \(p\)\\
\midrule
3B & GRPO     & 52 & 71.2\% & 0.0016\\
3B & DAPO     & 45 & 73.3\% & 0.0012\\
3B & Clip-Cov & 56 & 80.4\% & \(2.7{\times}10^{-6}\)\\
3B & KL-Cov   & 57 & 70.2\% & 0.0016\\
3B & EP-GRPO  & 52 & 73.1\% & 0.0006\\
3B & GTPO     & 42 & 50.0\% & 0.5612\\
\midrule
7B & GRPO     & 52 & 73.1\% & 0.0006\\
7B & DAPO     & 44 & 65.9\% & 0.0244\\
7B & Clip-Cov & 51 & 68.6\% & 0.0055\\
7B & KL-Cov   & 56 & 76.8\% & \(3.7{\times}10^{-5}\)\\
7B & EP-GRPO  & 57 & 82.5\% & \(<10^{-6}\)\\
7B & GTPO     & 40 & 57.5\% & 0.2148\\
\bottomrule
\end{tabular*}
\caption{Question-level direction check on the common-outcome subset.
``Failure higher'' is the fraction with
\(\widehat{\mathcal H}_{\mathrm{tok},-}(x)>
\widehat{\mathcal H}_{\mathrm{tok},+}(x)\); \(p\) is the one-sided exact
sign-test probability under an equal-direction null.}
\label{tab:sup-reward-sign}
\end{table}

The paired ratios in Table~\ref{tab:sup-reward-entropy} could in principle be
driven by a few large differences.  The direction check in
Table~\ref{tab:sup-reward-sign} instead gives every eligible question one
vote.  Failed entropy is higher for 65.9--82.5\% of questions and the
one-sided sign test rejects equal directions at ten of the twelve endpoints.
The two exceptions are GTPO, matching the paired-bootstrap intervals that
overlap one.  Thus, the cross-question result strengthens the descriptive
reward ordering without turning it into a causal statement.

\subsection{Mode-Retention Uncertainty}

\begin{table}[t]
\centering
\scriptsize
\setlength{\tabcolsep}{3pt}
\begin{tabular*}{\columnwidth}{@{\extracolsep{\fill}}llrrr@{}}
\toprule
Scale & Method & Ret. & 95\% CI & \(n\)\\
\midrule
3B & GRPO     & 94.2\% & [89.2, 99.9] & 72\\
3B & DAPO     & 91.1\% & [85.6, 96.7] & 72\\
3B & Clip-Cov & 96.0\% & [90.1, 102.5] & 73\\
3B & KL-Cov   & 94.8\% & [89.1, 100.6] & 72\\
3B & EP-GRPO  & 99.9\% & [92.3, 108.1] & 72\\
3B & GTPO     & 83.7\% & [77.4, 89.9] & 67\\
\midrule
7B & GRPO     & 89.6\% & [84.7, 94.5] & 81\\
7B & DAPO     & 88.8\% & [83.7, 93.9] & 80\\
7B & Clip-Cov & 93.5\% & [87.2, 100.7] & 82\\
7B & KL-Cov   & 92.3\% & [87.7, 96.9] & 81\\
7B & EP-GRPO  & 96.2\% & [89.5, 104.1] & 81\\
7B & GTPO     & 85.9\% & [79.9, 92.0] & 77\\
\bottomrule
\end{tabular*}
\caption{Question-bootstrap uncertainty for the published rarefied
effective-mode retention.  Intervals use 10,000 resamples and condition on
the frozen rarefaction draws and the single training seed.  Clip-Cov-7B uses
step 2,000; other rows use final checkpoints.}
\label{tab:sup-mode-bootstrap}
\end{table}

All twelve point estimates are below the base-policy reference of 100\%.
Seven question-bootstrap intervals lie wholly below 100\%; the other five
include equality because the question-level retention ratios are
heterogeneous.  The result supports the aggregate contraction reported in
the main paper, but not the stronger claim that every individual objective
has independently significant contraction.  These intervals quantify
question-sampling uncertainty only and do not substitute for additional
training seeds.

\subsection{Exact Covariance Fit}

\begin{table}[t]
\centering
\small
\setlength{\tabcolsep}{3.1pt}
\begin{tabular*}{\columnwidth}{@{\extracolsep{\fill}}lrrrr}
\toprule
Scope & \(n\) & Sign & \(\rho_s\) & \(r\)\\
\midrule
All nonfailed & 44 & 95.5\% & 0.986 & 0.968\\
3B & 24 & 95.8\% & 0.990 & 0.994\\
7B nonfailed & 20 & 95.0\% & 0.937 & 0.962\\
\bottomrule
\end{tabular*}
\caption{Finite-difference covariance fit by scale.  \(n\) counts adjacent
50-raw-step transitions; \(\rho_s\) and \(r\) are Spearman and Pearson
correlations.}
\label{tab:sup-cov-fit}
\end{table}

Pooling the 44 nonfailed transitions gives slope 0.993, intercept
\(4.37\times10^{-4}\), \(R^2=0.937\), median absolute error
\(1.18\times10^{-4}\), and median relative error 12.7\%.  The 3B and
nonfailed-7B slopes are 0.848 and 1.153, respectively.  The pooled near-unit
slope supports the local approximation, while the scale-specific deviations
and finite interval caution against interpreting it as an exact optimizer
identity.

\subsection{Covariance-Tail Sensitivity}

\begin{table}[t]
\centering
\small
\setlength{\tabcolsep}{3.1pt}
\begin{tabular*}{\columnwidth}{@{\extracolsep{\fill}}lrrrr@{}}
\toprule
Upper tokens & Success mass & Failure mass & Gap & Direction\\
\midrule
5.0\% & 25.3\% & 94.6\% & 69.3 pp & 12/12\\
2.5\% & 13.9\% & 94.2\% & 80.3 pp & 12/12\\
1.0\% & 5.6\% & 86.6\% & 81.0 pp & 12/12\\
\bottomrule
\end{tabular*}
\caption{Sensitivity to the rank-defined upper-tail budget.  Mass columns
macro-average the fraction of positive chosen-token covariance in each
reward stratum; ``Direction'' counts endpoints with greater failed-stratum
concentration.}
\label{tab:sup-cov-sensitivity}
\end{table}

The q99 statistic in the main paper retains cutoff ties, which is important
for reporting its literal empirical-quantile definition but can make the
selected token fraction large when covariance values are tied.  As a
tie-robust complement, Table~\ref{tab:sup-cov-sensitivity} fixes an
equal-count rank budget at the upper 1\%, 2.5\%, and 5\% of tokens.  At every
budget and all twelve comparative endpoints, failed prefixes allocate more
positive covariance mass to the selected tail.  The mean failed-minus-success
gap remains 69.3--81.0 percentage points.  The reward-stratified tail
concentration is therefore not an artifact of choosing q99 or of the GTPO
ties noted in the main analysis.

\subsection{Clip-Cov-7B Failure Diagnosis}

The failure detector is applied to every evaluated checkpoint in the dense
MATH500 trace; the plotted cross-scale curve uses the 20-point common-progress
grid.  The detector requires a peak accuracy of at least 30\%, an endpoint
accuracy at most 5\%, and a peak-to-endpoint drop of at least 30 percentage
points.  Only
Clip-Cov-7B triggers the rule: the full dense evaluation attains 76.6\% at
raw steps 1,000 and 2,400, with the last maximum at step 2,400 (30\%
progress); it reaches 0.2\% at step 4,000 and ends at 0.2\% at step 8,000.

\begin{table}[t]
\centering
\small
\setlength{\tabcolsep}{3.8pt}
\begin{tabular*}{\columnwidth}{@{\extracolsep{\fill}}rrrr}
\toprule
Raw step & Accuracy & Boxed rate & Mean length\\
\midrule
2,400 & 76.6\% & 99.6\% & 598.5\\
3,600 & 60.8\% & 95.2\% & 431.5\\
4,000 & 0.2\% & 1.0\% & 1,999.8\\
8,000 & 0.2\% & 0.4\% & 2,040.0\\
\bottomrule
\end{tabular*}
\caption{Dense MATH500 diagnostics for Clip-Cov-7B.  The generation limit is
2,048 tokens.}
\label{tab:sup-clip-failure}
\end{table}

The discontinuity is therefore a termination/length degeneration rather than
a gradual loss of mathematical accuracy.  In the surrounding 45--55\%
training window, only 21.0\% of sampled completions belong to prompt groups
with nonzero reward variance, and the covariance mask selects no token on
93.0\% of updates.  A plausible post-hoc explanation is that EOS/length drift
entered a self-reinforcing near-all-failure regime: the sparse interval-based
Clip-Cov intervention was usually inactive, while uniformly failed groups
provided no recovery gradient.  This diagnosis is observational and should
not be read as a causal ablation.

\section{Interpretation Boundaries}

The main paper's strongest empirical conclusions are the instability of
global-entropy performance rankings, the reward-conditioned endpoint
ordering, and the short-interval covariance fit.  The q99 result is a
reward-stratified endpoint concentration proxy.  Mode retention is an audited
but coverage-dependent diagnostic, and its association with MATH500 pass@64
is method-level, cross-dataset, single-seed, and exploratory.  The experiments
do not estimate within-mode realization entropy, mode-stratified covariance,
out-of-distribution robustness, or the causal effect of preserving a
particular mode.  Consequently, the paper reports continuous mode contraction
and does not assign formal harmful-collapse labels.

\bibliography{aaai2027}